\documentclass[conference]{IEEEtran}
\IEEEoverridecommandlockouts
\usepackage{cite}

\usepackage[table]{xcolor}
\definecolor{frozenrow}{HTML}{EAF2FA}

\usepackage{amsmath,amssymb,amsfonts}
\usepackage{algorithmic}
\usepackage{graphicx}
\usepackage{textcomp}
\usepackage{url}
\usepackage{booktabs}
\usepackage{array}
\usepackage{tikz}
\usetikzlibrary{arrows.meta}
\usepackage{caption}
\definecolor{cblue}{HTML}{2A78D6}
\definecolor{corange}{HTML}{EB6834}
\definecolor{cgray}{HTML}{7F7F7F}
\def\BibTeX{{\rm B\kern-.05em{\sc i\kern-.025em b}\kern-.08em
    T\kern-.1667em\lower.7ex\hbox{E}\kern-.125emX}}
\begin{document}

\title{Severe but Harmless: Stress-Testing Shift Severity for Frozen Time-Series Forecasters}

\author{\IEEEauthorblockN{Korede R. Bishi, Dominic Okonkwo, and Adetayo Okunoye}
\IEEEauthorblockA{\textit{School of Computing},
\textit{University of Georgia, Boyd Research and Education Center}\\
Athens, GA 30602, USA \\
korede.bishi, dominic.okonkwo, adetayo.okunoye@uga.edu}
}

\maketitle
\footnotetext{Code and result: \url{https://github.com/koredebishi/ST-TSFM}}

\begin{abstract}

Zero-shot time-series foundation models (TSFMs) have become attractive 
alternatives to task-specific models due to their ability to handle various 
downstream tasks in anomaly detection and forecasting across diverse time-series 
settings. However, time-series data are known to undergo distribution shifts, 
and the dissimilarity between two regime datasets is commonly measured using 
statistical measures such as Wasserstein-1, with higher scores interpreted as 
greater shift severity. In systematic experiments using three established severity 
measures on nine regime shifts of varying severity from three standard traffic
datasets, we show that these measures do not reliably predict degradation 
for instance-normalized TSFMs. For example, our results show that the regimes 
rated by these measures as the most severe are not the regimes that degrade the 
models the most, with degradation reaching up to $29\%$ in the lowest-scoring regime. 
In most cases, the audit results show no positive relationship between measured 
severity and model degradation, supporting our finding that measured severity 
does not reliably predict degradation for these TSFMs. We provide an 
equivariance-based stress-testing analysis that explains why severity 
readings can become decoupled from model degradation.

\end{abstract}

\vspace{1em}

\section{Introduction}

Time-series foundation models (TSFMs) such as Chronos-2~\cite{chronos2},
TimesFM~\cite{timesfm}, and Moirai~\cite{moirai} forecast new series from
recent history without training on each series, which is called zero-shot
forecasting. This makes them attractive alternatives to supervised models
built for individual tasks~\cite{yanespulido,tsfmpower}. But a series
encountered in use may differ from the data a model was trained on.
Weekday and weekend traffic, for example, follow different patterns. Such
differences are called distribution
shifts~\cite{gama2014,failingloudly,uncertaintyshift}; when a shift
develops over time, it is often called drift. A shift can raise forecast
error even when the data come from the same source. The problem extends
beyond time series to vision and tabular
data~\cite{wilds,doc,maggio,atc,aotl}. In time-series forecasting, several
methods address it by normalizing the data before
prediction~\cite{revin,nonstattransf,dishts}.

In traffic, these supervised models are graph
forecasters~\cite{dcrnn,gwnet,staeformer}, and on our three datasets they
still have lower error than the TSFMs (Section~\ref{sec:indist}). Although
distribution shift is widely studied in the literature, including for time
series, we did not find a study that audits distribution shift for TSFMs in
the traffic domain, characterizing the severity of these shifts and the
measures used to score it.

Wasserstein-1 is commonly used to measure how far the distributions of two
datasets differ~\cite{chenwass}. Traffic patterns can differ even within the
same dataset. Morning rush and midday on the same weekday, for example, may
have different distributions of readings. We call each condition used to
group observations a regime, including weekdays and weekends, peak and
off-peak periods, and groups of sensors. In principle, scoring regimes could
help prioritize the highest-scoring shifts for
investigation~\cite{gama2014,drifttools,driftdlrep,driftbench,peyrecuturi}
and guide whether similar regimes should be combined or kept separate.

In the literature, various methods have normalized the data before
prediction~\cite{revin,nonstattransf,dishts}. The advantage this gives is that
dissimilarities are first dealt with in the data so that the models can better
learn good representations from them. TSFMs already do this: they
standardize every context window by its own mean and standard deviation
before predicting~\cite{chronos2}. The problem is that a distance computed on raw
readings mostly measures level and scale, which is exactly what the model
removes. Suppose we take a weekend that halves traffic volume across a network.
The distance sees a large shift and Wasserstein-1 will report this;
however, after standardizing, the TSFM sees an input much like its usual one. The
severity measure and the model are not looking at the same thing, and once that is
true, severity can move without the model's error moving with it.

Our goal in this work is therefore to investigate whether these marginal
severity measures (Wasserstein-1, Kolmogorov--Smirnov, and mean
difference) can reliably identify the distribution shifts that degrade
Chronos-2~\cite{chronos2} and TimesFM~\cite{timesfm}. We use the
released models frozen, without any training.

\begin{figure*}[t]
\centerline{\includegraphics[width=\textwidth]{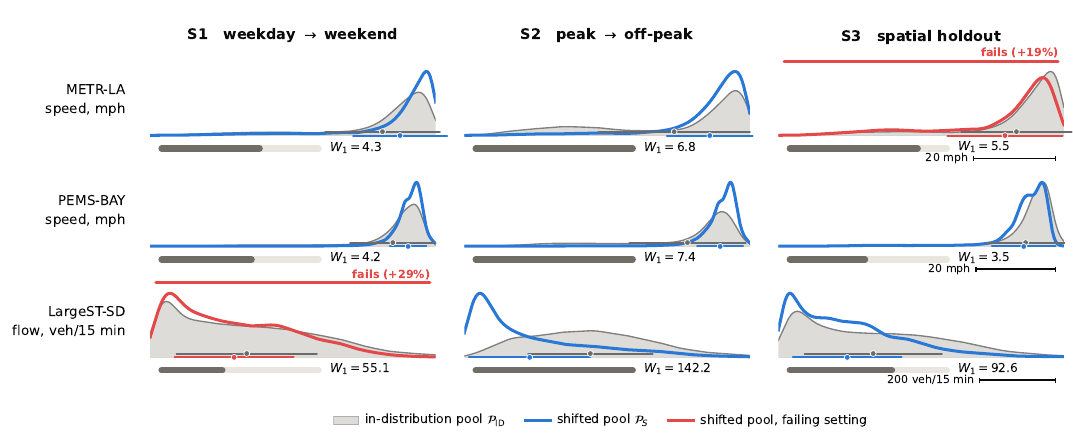}}
\caption{The nine dataset--shift settings (gray: reference, lines: shifted).
The temporal failure, LargeST-SD S1 (red), has the lowest severity in its row.
Bars show normalized $W_1$.}
\label{fig}
\end{figure*}

We summarize our contributions below:

\begin{itemize}
\item We test whether three marginal severity measures,
Wasserstein-1, Kolmogorov--Smirnov, and mean difference, identify the
distribution shifts that degrade Chronos-2 and
TimesFM by evaluating nine dataset--shift settings
across three real-world traffic datasets against five baselines.

\item We show that severity and degradation can become decoupled for
instance-normalized forecasters because per-window normalization removes
the level and scale changes that can dominate marginal severity. We
verify this affine-equivariance property in the released Chronos-2 and
TimesFM checkpoints.

\item We show empirically that the severity measures do not identify
the degrading regimes: the highest-severity shifts are largely
non-degrading, while LargeST-SD weekends produce up to $29\%$
degradation at the lowest severity reading in that dataset. We also
report a pre-registered test-time adaptation experiment that fails to
repair this weekend degradation.

\item We show that a validation-selected timestamp router outperforms
both frozen models on LargeST-SD weekends, reaching a mean absolute
error (MAE, lower is better) of 21.57 veh/15\,min compared
to 23.82 for TimesFM and 26.14 for Chronos-2, though trained graph
models remain stronger there.
\end{itemize}

\section{Related Work}

\subsection{Time-Series Foundation Models}

Pretrained forecasters now transfer to unseen series without
target-domain training. Chronos discretizes series into a
language-modeling vocabulary~\cite{chronos}, TimesFM trains a
decoder-only forecaster on a large corpus~\cite{timesfm}, and Moirai and MOMENT extend this model class with universal multivariate and open
masked-reconstruction pretraining~\cite{moirai,moment}. Chronos-2 standardizes each context window internally~\cite{chronos2}. This normalization is central to the analysis in Section~\ref{sec:decouple}. Deployment results differ between aggregate and regime-level evaluation. Benchmarked across ten transportation datasets, zero-shot Chronos-2 is
state-of-the-art or competitive on aggregate error~\cite{yanespulido}, yet zero-shot foundation forecasters fail on cloud workloads~\cite{tonercloud},
and regime-stratified scoring of traffic speed finds 3~mph aggregate
errors concealing transition errors near 11~mph~\cite{wangregime}. Aggregate evaluation is further limited by contamination: standard
benchmarks appear verbatim in public pretraining
corpora~\cite{moirai,meyerleakage}, so a TSFM may have seen the test
data. Tools to audit this leakage are only emerging~\cite{tsfmaudit}. These results therefore do not establish reliability at the regime level required for
deployment.

\subsection{Distribution Shift in Spatio-Temporal Forecasting}
Traffic forecasting has a mature literature on supervised graph and
transformer forecasters~\cite{dcrnn,gwnet,staeformer} benchmarked at scale by LargeST~\cite{largest}. Shift studies with these models retrain on the new
regime and report the resulting performance gap, while
out-of-distribution benchmarks document degradation without
establishing a relationship between shift magnitude and
harm~\cite{woods}.

\subsection{Normalization and Affine Equivariance}
\label{sec:normlineage}

Traffic is nonstationary: its mean and variance change over time. A
standard architectural defense is normalization. Reversible instance normalization~\cite{revin}, series
stationarization~\cite{nonstattransf}, and learned per-window
statistics~\cite{dishts} remove each window's level and scale before
prediction, a design also used by the foundation models considered here~\cite{chronos2}.
In image denoising, networks with $f(ax+b)=af(x)+b$ for $a>0$ are
called normalization-equivariant. They are built layer by
layer~\cite{normequiv23}, or by normalizing the input and reversing it on
the output, which gives exactly this class~\cite{normequiv26}.
Proposition~1 applies this property to forecasters. Those works use it
to design models. We use it to explain why a raw-data distance can
weigh heavily what the model removes before prediction.

\subsection{Severity Metrics in Audit Practice}
Deployment monitoring often begins by comparing incoming data with a
reference distribution. Concept-drift research studies how to detect such
changes and adapt to them~\cite{gama2014}. Practical approaches include
drift detectors, representation-based monitors, and drift
benchmarks~\cite{drifttools,driftdlrep,driftbench}. Optimal transport
provides one way to measure the difference~\cite{peyrecuturi}.

A detected shift does not necessarily predict a model's error. Some shifts
are harmless~\cite{failingloudly}, and uncertainty estimates can become
unreliable under shift~\cite{uncertaintyshift}. Studies of shifts in the
wild also find that magnitude does not consistently track
loss~\cite{wilds}. Similar limits appear in performance estimation.
Fr\'echet and MMD estimates have been unreliable in vision~\cite{doc},
shift-detection scores have been poor predictors on tabular
data~\cite{maggio}, and confidence-based estimates can outperform
distance-based ones~\cite{atc}. Out-of-distribution accuracy can follow
in-distribution accuracy~\cite{aotl}, but that prediction uses the
model's own performance rather than a distance between raw marginals.

For time series, one cross-dataset study reports a correlation above 0.60
between Wasserstein distance and foundation-model loss~\cite{chenwass}.
That result does not establish whether distance also ranks the effect of
shifts among operating regimes within one deployment. This is the
assumption we test here.

\subsection{Test-Time Adaptation for Time Series}

Test-time adaptation updates a deployed model using unlabeled inputs,
via self-supervised objectives in
vision~\cite{tttsun} and, for forecasting, through frequency
calibration, lightweight input--output modules, and feature
adjustment~\cite{princtta,peft_tta,onlinefeatadj}. Other studies document cases in which adaptation degrades performance: effectiveness
is sensitive to shift type and model choice~\cite{ttapitfalls}, continual adaptation can perform worse than the non-adapted model~\cite{rdumb}, and performance deteriorates on temporally correlated streams~\cite{note_tta}. Adaptation under time-series
dependence is therefore an empirical question rather than a
guaranteed repair.

\section{Methodology}
Our approach is to audit the severity measure under the conditions of its
use: fixed pools, one masking convention, and a protocol fixed before the
shift evaluations. We registered the predictions in advance and report them
with their outcomes in Section~\ref{sec:preds}.

 \subsection{Problem Setup and Notation}
 Each model uses the recent history of every sensor to forecast the
 next 12 steps. METR-LA (207 sensors) and PEMS-BAY record speed every
 5 minutes, so the 3rd, 6th, and 12th forecast steps (h3, h6, h12)
 are 15, 30, and 60 minutes ahead,
 following the standard traffic-forecasting convention~\cite{dcrnn}.
 LargeST-SD records flow every 15 minutes, making the same horizons
 45, 90, and 180 minutes.

 We describe all three datasets using the same notation. Let
 $y_i(t)$ denote the reading from sensor
 $i \in \{1,\dots,N\}$ sampled at cadence $\delta$. A window $w$
 anchored at time $t_w$ pairs a context
 $X_w = \{y_i(t)\}_{t_w - C < t \le t_w}$ of $C = 512$ steps
 (equal for both models, a multiple of TimesFM's 32-step patch)
 with targets $Y_w \in \mathbb{R}^{N \times H}$, $H = 12$, and a
 forecaster $f$ maps $X_w$ to $\hat{Y}_w$. We score forecasts by mean
 absolute error (MAE; lower is better), masked and averaged over a set
 of windows, or pool, $\mathcal{P}$,
\begin{equation}
\mathrm{MAE}(f;\mathcal{P}) \;=\; \frac{1}{|V(\mathcal{P})|}
\sum_{(w,i,h) \in V(\mathcal{P})}
\bigl|\hat{Y}_{w,i,h} - Y_{w,i,h}\bigr|,
\label{eq:mae}
\end{equation}
where
$V(\mathcal{P}) = \{(w,i,h) : w \in \mathcal{P},\, Y_{w,i,h} \neq 0\}$
excludes the zero-coded missing entries identically for every model,
the masking convention established by~\cite{dcrnn} and standard across
the traffic literature~\cite{gwnet,largest}.

\subsection{Shift Construction}

We construct three shifts per dataset. S1 (weekday to weekend) and S2
(peak to off-peak) are temporal: weekdays are Monday through Friday,
and peak hours are 7--10 and 16--19 local time
(Section~\ref{sec:ablation} varies this window). For each temporal
shift, $\pi_S$ is a condition on the timestamp (for S1, ``Saturday or
Sunday''), and a test window enters the shift pool only if all $H$
steps it forecasts satisfy it,
\begin{equation}
\mathcal{P}_S \;=\; \bigl\{\, w \;:\; \pi_S(t_w + h\delta)\
\text{holds for all}\ h = 1,\dots,H \,\bigr\}.
\label{eq:pool}
\end{equation}
Requiring the full forecast horizon to satisfy the condition prevents
windows that cross a regime boundary from entering the pool. S3 (spatial
holdout) instead tests sensors in a different location. We
cluster sensor coordinates with $k$-means ($k=5$, fixed seed;
Section~\ref{sec:ablation} tests $k \in \{4,6,8\}$) and hold
out one cluster: 37 of 207 sensors on METR-LA, 36 of 325 on PEMS-BAY,
and 102 of 716 on LargeST-SD. The S3 pool contains every test window
restricted to the held-out sensors, with the retained sensors forming
the in-distribution sensor set.

\subsection{Severity Measurement}

With the shift pools fixed, we measure severity as the Wasserstein-1
distance between the in-distribution and shifted marginals of the
observable, that is, the distributions of raw readings (speed or flow)
in each pool, ignoring time order. Following~\cite{peyrecuturi}, for
these scalar marginals the distance reduces to the closed quantile-function form
\begin{equation}
W_1(\mu,\nu) \;=\; \int_0^1
\bigl|F_\mu^{-1}(u) - F_\nu^{-1}(u)\bigr|\, du.
\label{eq:w1}
\end{equation}
We compute this distance per sensor and report the sensor mean. For
the two temporal shifts, S1 and S2, each sensor's in-distribution and
shifted target pools supply the two marginals. The spatial case differs
because S3 has no shared sensor axis: each held-out sensor's marginal
is compared with the pooled marginal of the retained sensors. S3 thus
compares a sensor with other sensors, not with itself, so its reading
also carries location differences. With all sensors pooled on both
sides instead, LargeST-SD weekends are still the lowest-severity
setting ($W_1$ 49.7, against 68.3 and 140.8).

For comparisons across settings, we normalize severity within each
dataset by its largest reading. Alongside $W_1$, we compute two further
marginal severity measures under the same conventions: the two-sample
Kolmogorov--Smirnov statistic and the absolute mean difference, which
isolates the level component. Section~IV reports all three.

Fig.~\ref{fig} shows the nine pairs of pool marginals used by these
estimators; the resulting readings are reported in
Table~\ref{tab1}.

\subsection{Models and Training Protocols}
\label{sec:zoo}

With the shifts and severity measures fixed, we evaluated seven models:
two frozen TSFMs (Chronos-2 and TimesFM),
three trained graph models (DCRNN, Graph WaveNet, and STAEformer), and
two classical baselines (historical average, HA, and ARIMA). The TSFMs operate zero-shot
and are never tuned. We train the graph models on the standard split
of each dataset in accordance with their released
configurations~\cite{dcrnn,gwnet,staeformer}. The LargeST-SD variants
of DCRNN and Graph WaveNet use the benchmark authors' released
implementations~\cite{largest}; STAEformer uses its upstream
implementation on converted SD data, with alignment checked against the
frozen arrays.

For the repair analysis of Section~\ref{sec:repair}, we use the same
graph architectures but retrain DCRNN and Graph WaveNet on the
corresponding in-distribution training pools. For S1, this means
weekday-only target windows; for S2, peak-only target windows. We then
evaluate these retrained models on the shifted pools using the same
masking convention.

\subsection{Evaluation Criteria}
We evaluate each model under three criteria. For model $f$ and shift
$S$ (S1, S2, or S3), the first measures relative MAE change,
\begin{equation}
\Delta_S(f) \;=\;
\frac{\mathrm{MAE}(f;\mathcal{P}_S)}
{\mathrm{MAE}(f;\mathcal{P}_{\mathrm{ID}})} - 1.
\label{eq:deg}
\end{equation}
A positive value means higher MAE under the shift.
Table~\ref{tab:deg} reports $100\Delta_S(f)$ in percent. The
in-distribution pool $\mathcal{P}_{\mathrm{ID}}$ is the complementary
slice of the same test period: weekday-target windows for S1,
peak-target windows for S2, and retained sensors for S3. The paired
pools therefore share the alignment and masking conventions.

The second criterion asks whether the shift changes which model is
preferred. We rank the $M$ models available on each dataset by MAE on
$\mathcal{P}_{\mathrm{ID}}$ and again on $\mathcal{P}_S$, and compare
the two orders with Kendall's rank correlation ($\tau=1$ means the
ranking does not change),
\begin{equation}
\tau = \frac{n_c - n_d}{\binom{M}{2}},
\label{eq:tau}
\end{equation}
where $n_c$ and $n_d$ are the numbers of concordant and discordant
model pairs. We use $M = 7$ for METR-LA and PEMS-BAY and $M = 6$ for
LargeST-SD, the model set fixed when the ranking protocol was
registered. The released artifacts also include the $M = 7$
LargeST-SD variant. We form confidence
intervals by resampling sensors with replacement using 1000 resamples
and a fixed seed. For S3, the two disjoint sensor sets are resampled
independently. The interval endpoints are the 2.5th and 97.5th
percentiles of the resulting $\tau$ distribution.

The third criterion asks whether a failing shift can be repaired. We
measure adaptability by whether an intervention improves upon the
corresponding frozen model. The test-time adaptation protocol for this
criterion is formalized in
Section~\ref{sec:case}.

Finally, to test whether severity orders the shifts by degradation
(the first criterion), we
compute the Spearman rank correlation between within-dataset normalized
severity and $\Delta_S$. We report this correlation at two granularities: descriptively over
all 18 model--setting evaluations, and inferentially at the setting
level, where the two backbones are averaged per setting and the exact
two-sided $p$-value is computed over all $9!$ orderings. For the
failing temporal setting we additionally form day-block bootstrap
intervals, resampling calendar days within each pool (2000 resamples,
fixed seed). We record model-specific freedoms, including context
length and batching, for every run.

\subsection{Theoretical Analysis: Severity--Degradation Decoupling}
\label{sec:decouple}

Chronos-2 first standardizes each context window by its own mean and
standard deviation. The network therefore forecasts in normalized
coordinates, after which the transformation is inverted to return to
the original scale. For the argument below, let $x$ be one sensor's
context window and $y$ its target. The full context $X_w$ in
Section~III-A stacks these series across sensors. What matters
is what this normalization does to a change in level or scale. The
answer is the normalization-equivariance of~\cite{normequiv23,normequiv26},
which we restate for forecasters because the consequence below rests on it.

\textit{Proposition 1 (affine equivariance).} Let $f$ be a forecaster
of the form
\[
f(x) = s(x)\, g\!\left(\frac{x - m(x)}{s(x)}\right) + m(x),
\]
where $m$ and $s$ are the context mean and standard deviation,
$s(x) > 0$, and $g$ is any map acting in the normalized domain. Then,
for every $a > 0$ and $b$, $f(ax + b) = a f(x) + b$.

\textit{Proof.} For $a > 0$, $m(ax+b) = a\,m(x) + b$ and
$s(ax+b) = a\,s(x)$. The standardized context
$z = (x - m(x))/s(x)$ is therefore invariant under $x \mapsto ax + b$.
Hence
\[
\begin{aligned}
f(ax+b) &= s(ax+b)\,g(z) + m(ax+b) \\
     &= a\,s(x)\,g(z) + a\,m(x) + b \;=\; a f(x) + b.
\end{aligned}
\]

\textit{Consequence.} The proposition has a direct consequence for a
severity audit. Suppose a regime differs only by a translation,
$(x+b,\,y+b)$. The model prediction moves by the same $b$, so the
forecast error does not change. Yet the marginal
Wasserstein-1 distance for that same shift is $|b|$, which can be made
arbitrarily large. The severity score can therefore keep growing
while degradation stays exactly zero. Scaling behaves similarly since
absolute error scales with the data, but the relative comparison is unchanged. In this way, marginal severity and degradation become
decoupled for this model class.

Chronos-2 also applies arcsinh to the standardized values, which
compresses extreme values, and sinh on the way back. Both act after
standardization, so they are part of $g$ and Proposition~1 still
holds. We checked the argument against the released
models rather than relying on the equation alone. For Chronos-2,
scaling a real window by $a=2$ gives bitwise-matching transformed
forecasts. With a much larger translation, $b=1000$, the MAE
ratio deviates from 1 by 0.78\%, which we attribute to
bfloat16 arithmetic. TimesFM normalizes with running per-patch mean and standard deviation
statistics, which are affine-equivariant in the same way; across the same
transformations its worst relative deviation is below $2\times10^{-7}$.
The verification scripts are in the released repository. This result is
limited to forecasters with this form of instance normalization.

\section{Experiments}

\subsection{Experimental Setup}
We use three real-world traffic datasets comprising 1,248 sensors and
roughly 49 million observations. METR-LA and PEMS-BAY record traffic
speed, while LargeST-SD records traffic flow.
Every model forecasts every frozen test window, which
amounts to about 60 million target values per model on LargeST-SD
alone. We use LargeST's released single-region year at full sensor
count. Frozen models produce one prediction array over the full test period per
dataset. We use the released Chronos-2 checkpoint and TimesFM~2.5
(200M parameters, PyTorch release). Both frozen models receive the same inputs: a context of 512
steps per sensor, batched 256 sensors per forward pass. Forecasts are
invariant to the ordering of sensors within a batch for both models,
so no cross-sensor information enters the frozen predictions.
Shift pools are formed from the same arrays by the target-step
rule, so in-distribution and shifted scores share every alignment
convention. The shift-retrained variants used in
Section~\ref{sec:repair} follow the protocol of Section~\ref{sec:zoo}.

Table~\ref{tab:cost} reports the one-time compute. Under our released
implementations, frozen-model inference accounts for most of this cost.

\begin{table}[htbp]
\centering
\renewcommand{\arraystretch}{1.15}
\footnotesize
\caption{One-time audit compute.}
\label{tab:cost}
\begin{tabular*}{\columnwidth}{@{\extracolsep{\fill}}llr@{}}
\toprule
\textbf{Task} & \textbf{Dataset} & \textbf{Wall time} \\
\midrule
Chronos-2 inference      & All 3 datasets & 0.7 h \\
TimesFM inference        & All 3 datasets & 192.8 h \\
Graph WaveNet training   & LargeST-SD     & 2.2 h \\
STAEformer training      & LargeST-SD     & 4.0 h \\
TTRA adapter             & LargeST-SD S1  & 2.8 h \\
Table/figure generation  & All results    & $\leq 33$ s \\
\bottomrule
\end{tabular*}
\\[2pt]
{\footnotesize Runs use an NVIDIA A100. Predictions are cached once;
all tables and figures use the cached results. Both models receive
identical inputs. With its released code, TimesFM~2.5 ran at about 15
series per second for any batch size from 256 to 4,096.}
\end{table}

\subsection{In-Distribution Performance}
\label{sec:indist}
Before evaluating distribution shift, we first compare the models on
the standard test split. The trained graph models achieve the lowest
MAE on all three datasets, while the frozen TSFMs generally fall
between the graph models and the classical baselines. On METR-LA,
Chronos-2 records 4.65 MAE, compared with 4.19 for the historical
average. The Std column of Table~\ref{tab:tbig} reports the full
comparison. We establish this ordering as a reference for the shift
results that follow.

\subsection{Finding 1: Severity Does Not Predict Degradation}
\label{sec:f1}
We first ask whether the shifts ranked as more severe are also the
shifts that degrade the frozen models. In 7 of 9 dataset--shift
settings (14 of 18 model--setting evaluations), the models maintain or
improve performance under shift. PEMS-BAY improves by as much as 54.7\%,
and the off-peak regimes are easier on all three datasets. The two
exceptions are LargeST-SD weekends (+29\% for Chronos-2, +8\% for
TimesFM) and the METR-LA spatial holdout (+19\%, +21\%).
Table~\ref{tab:deg} places these degradation values beside each
setting's severity reading, while Table~\ref{tab:tbig} reports the
corresponding MAE for all seven models.

\begin{table}[htbp]
\centering
\renewcommand{\arraystretch}{1.15}
\setlength{\tabcolsep}{1.3pt}
\footnotesize
\caption{Shift severity and frozen-model degradation.}
\label{tab1}\label{tab:deg}
\begin{tabular}{lccccccccc}
\toprule
& \multicolumn{3}{c}{\textbf{S1 (Weekend)}} &
  \multicolumn{3}{c}{\textbf{S2 (Off-peak)}} &
  \multicolumn{3}{c}{\textbf{S3 (Spatial)}} \\
\cmidrule(lr){2-4}\cmidrule(lr){5-7}\cmidrule(lr){8-10}
\textbf{Dataset} &
$W_1$ & $\Delta$C2 & $\Delta$TFM &
$W_1$ & $\Delta$C2 & $\Delta$TFM &
$W_1$ & $\Delta$C2 & $\Delta$TFM \\
\midrule
METR-LA
& \underline{4.3} & $-1.5$ & $-4.5$
& 6.8 & $-28.3$ & $-35.5$
& 5.5 & $\mathbf{+18.7}$ & $\mathbf{+20.6}$ \\

PEMS-BAY
& 4.2 & $-51.4$ & $-48.1$
& 7.4 & $-52.9$ & $-54.7$
& \underline{3.5} & $-19.9$ & $-20.1$ \\

LargeST-SD
& \underline{55.1} & $\mathbf{+28.6}$ & $\mathbf{+8.4}$
& 142.2 & $-33.7$ & $-35.2$
& 92.6 & $-26.1$ & $-25.7$ \\
\bottomrule
\end{tabular}\\[2pt]

{\footnotesize $W_1$ is Wasserstein-1 in observable units (mph;
veh/15\,min for SD), so compare $W_1$ only within a row. Underline
marks the lowest $W_1$ in each row. $\Delta$C2 and $\Delta$TFM report
$100\Delta_S(f)$ for Chronos-2 and TimesFM; bold marks degradation.}
\end{table}
\begin{table*}[t]
\centering
\renewcommand{\arraystretch}{1.15}
\setlength{\tabcolsep}{4.5pt}
\footnotesize
\caption{Average MAE on the standard test split and each shift pool.}
\label{tab:tbig}
\begin{tabular*}{\textwidth}{@{\extracolsep{\fill}}lcccccccccccc}
\toprule
& \multicolumn{4}{c}{\textbf{METR-LA (mph)}} &
  \multicolumn{4}{c}{\textbf{PEMS-BAY (mph)}} &
  \multicolumn{4}{c}{\textbf{LargeST-SD (veh/15\,min)}} \\
\cmidrule(lr){2-5}\cmidrule(lr){6-9}\cmidrule(lr){10-13}
\textbf{Model}
& Std & S1 & S2 & S3
& Std & S1 & S2 & S3
& Std & S1 & S2 & S3 \\
\midrule
HA
& 4.19 & 3.52 & 3.44 & 4.99
& 2.63 & 1.71 & 1.94 & 2.26
& 27.93 & 21.57 & 20.16 & 22.32 \\

ARIMA
& 5.23 & 5.45 & 4.77 & 5.87
& 2.23 & 1.28 & 1.76 & 1.84
& 55.43 & 42.16 & 45.47 & 42.62 \\

DCRNN
& 3.16 & 2.73 & 2.71 & 3.73
& 1.61 & 1.02 & 1.23 & 1.34
& 21.20 & 17.73 & 16.31 & 16.10 \\

Graph WaveNet
& 3.04 & 2.52 & 2.64 & 3.59
& 1.58 & 1.00 & 1.22 & 1.31
& 18.50 & 16.21 & 14.19 & 14.64 \\

STAEformer
& 2.94 & 2.47 & 2.58 & 3.45
& 1.54 & 0.98 & 1.18 & 1.24
& 18.44 & 16.17 & 13.65 & 14.34 \\

\rowcolor{frozenrow}
Chronos-2
& 4.65 & 4.61 & 4.10 & \textbf{5.34}
& 1.84 & 1.05 & 1.37 & 1.51
& 21.70 & \textbf{26.14} & 17.30 & 16.67 \\

\rowcolor{frozenrow}
TimesFM
& 3.91 & 3.78 & 3.38 & \textbf{4.56}
& 1.69 & 1.02 & 1.24 & 1.38
& 22.20 & \textbf{23.82} & 17.30 & 17.13 \\

\bottomrule
\end{tabular*}\\[2pt]

{\footnotesize Std = standard test split; S1 = weekends; S2 =
off-peak; S3 = held-out sensors. Bold marks the two settings where
Chronos-2 and TimesFM degrade. For S3, the trained models saw all
sensors during training.}
\end{table*}

The severity readings do not follow this degradation pattern.
LargeST-SD weekends, the strongest temporal failure, have the lowest
$W_1$ in that dataset (55.1, against 92.6 and 142.2), while the METR-LA spatial failure has a
mid-range reading. The highest-severity settings are instead the
off-peak regimes, where both frozen models improve. Across all 18
evaluations, within-dataset normalized severity and degradation have a
descriptive Spearman correlation of $\rho=-0.51$. Averaging the two
backbones within each setting gives $\rho=-0.47$ with $p=0.20$.
The same pattern appears with KS and mean difference
($\rho=-0.24$ and $-0.44$). Mean difference also reproduces the
Wasserstein within-dataset ordering, showing that $W_1$ on these pools
largely follows level change, the component removed by instance
normalization. Computing $W_1$ and KS on per-window standardized data,
as the models see it, does not help ($\rho=-0.56$ and $-0.44$).

\begin{figure*}[t]
\centerline{\includegraphics[width=\textwidth]{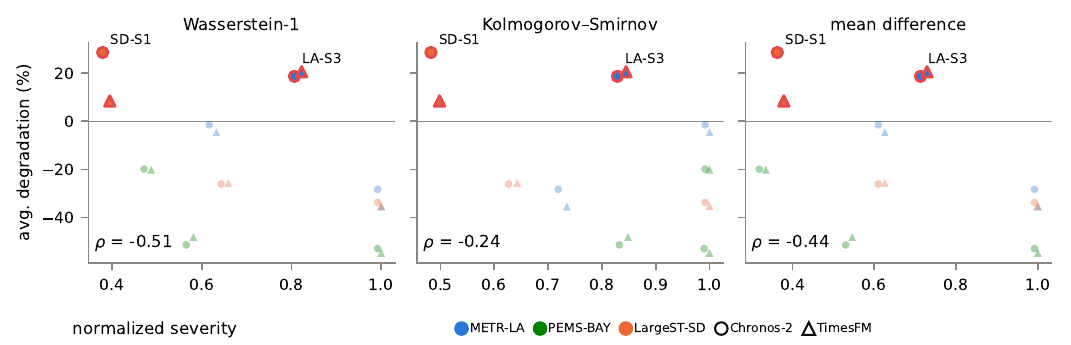}}
\caption{Severity versus degradation under $W_1$, KS and mean difference.
The temporal failure (SD-S1, red) has the lowest severity in its dataset under all
three. Panel $\rho$ is descriptive (18 non-independent points); exact
setting-level tests give $\rho=-0.47,-0.25,-0.41$ ($p=0.20,0.51,0.28$).}
\label{fig:scatter}
\end{figure*}

One possibility is that this negative association reflects
differences in regime difficulty. We therefore repeat the comparison
using loss of skill relative to the historical average. The
correlation disappears ($\rho=+0.02$, $p=0.98$), but LargeST-SD
weekends still show the largest skill collapse of all nine settings
(0.56 for Chronos-2 and 0.40 for TimesFM) at the lowest severity
reading in that dataset. The METR-LA spatial setting no longer counts
as a failure under this metric because the frozen models lose less
accuracy than the historical average. With MASE the correlation stays
negative ($\rho=-0.51$), but with sMAPE it is near zero
($\rho=-0.12$), because LargeST-SD off-peak, where flow is low, worsens
by 63\% in relative error for Chronos-2. Fig.~\ref{fig:scatter} shows all
18 evaluations under the three severity measures.

The LargeST-SD spatial result is split-sensitive: alternative holdout
cluster counts reverse its sign and reduce the non-degrading count from
7 of 9 to 6 of 9 (Section~\ref{sec:ablation}). No other result in this
finding depends on that spatial sign.
\subsection{Finding 2: Severity Misses a Ranking Failure}
\label{sec:f2}

\begin{figure}[t]
\centerline{\includegraphics[width=\columnwidth]{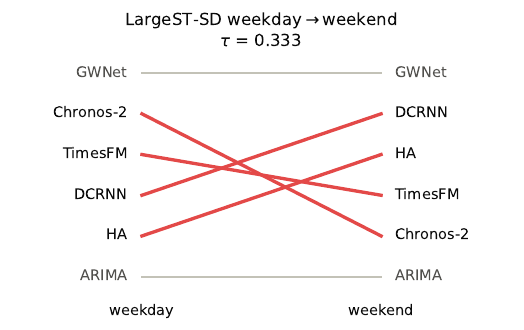}}
\caption{LargeST-SD model rankings on weekdays and weekends
($\tau=0.333$). The historical average moves above Chronos-2 and
TimesFM, while Chronos-2 falls from second to fifth.}
\label{fig:slopes}
\end{figure}

Having shown that severity does not identify the degrading settings,
we next ask whether those failures change model selection. Rankings
remain stable under most shifts: Kendall $\tau$ is at least 0.905 on
METR-LA and 1.000 on PEMS-BAY. LargeST-SD weekends are the exception,
where $\tau$ falls to 0.333 and the historical average moves above
both frozen TSFMs. A frozen model selected on weekday validation is
therefore not necessarily the best choice for the weekend regime.
Table~\ref{tab:tau} reports the full ranking comparison, and
Fig.~\ref{fig:slopes} shows this reordering directly.

Smaller pairwise inversions also occur on METR-LA, where the historical
average passes TimesFM on weekends and Chronos-2 off-peak, but the
overall ranking remains close to its in-distribution order. The
LargeST-SD result also persists when calendar days are resampled:
day-block bootstrap gives $\tau\in[0.20,0.47]$. For degradation,
Chronos-2 has an interval of $[13\%, 45\%]$, which excludes zero,
while TimesFM's $[-2\%, 20\%]$ interval does not.
\begin{table}[htbp]
\centering
\renewcommand{\arraystretch}{1.15}
\footnotesize
\caption{Agreement between in-distribution and shifted model rankings
(Kendall $\tau$).}
\begin{tabular*}{\columnwidth}{@{\extracolsep{\fill}}lccc}
\toprule
\textbf{Dataset} &
\textbf{S1 (Weekend)} &
\textbf{S2 (Off-peak)} &
\textbf{S3 (Spatial)} \\
\midrule
METR-LA & 0.905 & 0.905 & 1.000 \\
PEMS-BAY & 1.000 & 1.000 & 1.000 \\
LargeST-SD & $\mathbf{0.333}$ & 0.867 & 1.000 \\
\bottomrule
\end{tabular*}\\[2pt]

{\footnotesize Bold marks LargeST-SD weekends. Its 95\% confidence
interval is $[0.333,0.333]$ by sensor (1000 resamples; the ranking
never changed under sensor resampling) and
$[0.20,0.47]$ by day (2000 resamples).}

\label{tab:tau}
\end{table}

\subsection{Finding 3: Repairing the LargeST-SD Weekend Failure}
\label{sec:repair}

\begin{figure}[t]
\centerline{\includegraphics[width=\columnwidth]{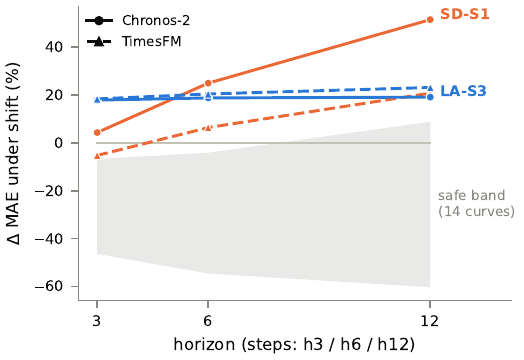}}
\caption{Degradation by forecast horizon. On LargeST-SD weekends it grows
to $+51\%$ for Chronos-2 at h12; on LA-S3 it stays near $+20\%$.
Shading spans the other 14 curves.}
\label{fig:deghz}
\end{figure}

We focus our repair analysis on LargeST-SD weekends because it is the only failing
temporal setting and the only weekend regime that is harder than its
weekday reference. It is a Saturday failure: Chronos-2 degrades by
67\% on Saturdays and 2\% on Sundays (TimesFM: 34\% and $-10\%$). This failure also becomes stronger with forecast
horizon, reaching $+51\%$ for Chronos-2 at h12. For METR-LA, its spatial
failure shows the same horizon-dependent pattern, while no
non-degrading setting exceeds $9\%$ at any horizon
(Fig.~\ref{fig:deghz}).

We first test retraining. If the failure can be corrected
from the reference regime alone, a model retrained on weekdays should
recover on weekends. It does not. Weekday-retrained DCRNN reaches
$28.96 \pm 0.72$ MAE on LargeST-SD weekends, worse than frozen
Chronos-2 at 26.14. Weekday-retrained Graph WaveNet reaches
$24.67 \pm 1.79$ and does not reliably improve on frozen TimesFM at
23.82. We also notice a broader pattern across the five settings with
both retrained and frozen results. The frozen models have lower MAE in
four, with METR-LA weekends the only exception.
The above findings leave two possibilities: either the retrained models have
too little data or are missing the regime on which they are evaluated.
We then separate these two possibilities at the same training budget. 
A random standard subset that includes weekends reaches $15.35 \pm 0.27$ MAE, while weekend-only Graph WaveNet reaches $15.60 \pm 0.13$ and weekend-only
DCRNN reaches $19.28 \pm 1.55$. In contrast, a weekday subset
size-matched to the weekend pool reaches $29.27 \pm 2.63$. The
failure therefore follows regime exclusion rather than training
volume.

\begin{table}[htbp]
\centering
\renewcommand{\arraystretch}{1.15}
\footnotesize
\caption{Model performance on LargeST-SD weekends.}
\begin{tabular*}{\columnwidth}{@{\extracolsep{\fill}}llc}
\toprule
\textbf{Model} & \textbf{Intervention} & \textbf{Avg MAE} \\
\midrule
Chronos-2 & None & 26.14 \\
TimesFM & None & $\mathbf{23.82}$ \\
Chronos-2 & TTRA adaptation & 27.80 \\
Chronos-2 & 48-hour context & 26.80 \\
Chronos-2 & 24-hour context & 29.76 \\
TimesFM & 48-hour context & 23.11 \\
TimesFM & 24-hour context & 33.15 \\
Graph WaveNet & Weekday-only retraining & $24.67 \pm 1.79$ \\
Graph WaveNet & Weekend-only training & $15.60 \pm 0.13$ \\
Graph WaveNet & Standard subset, size-matched & $15.35 \pm 0.27$ \\
Graph WaveNet & Weekday subset, size-matched & $29.27 \pm 2.63$ \\
Graph WaveNet & Per-window normalization & $27.85 \pm 0.61$ \\
DCRNN & Weekday-only retraining & $28.96 \pm 0.72$ \\
DCRNN & Weekend-only training & $19.28 \pm 1.55$ \\
Regime router & Timestamp routing & $\mathbf{21.57}$ \\
\bottomrule
\end{tabular*}\\[2pt]
{\footnotesize Retrained models report mean $\pm$ std over three seeds.
Bold marks the best frozen model and the router.}
\label{tab:retrain}
\end{table}

The standard-split graph models support the same explanation. Because
their training data already contain weekends, Graph WaveNet reaches
16.21 MAE and DCRNN 17.73 on the identical weekend pool
(Table~\ref{tab:tbig}). These results point to regime exposure as the main difference: models trained with weekend data perform well on the
weekend pool, whereas models trained only on weekdays do not.

\label{sec:case}

\begin{figure}[t]
\centerline{\includegraphics[width=\columnwidth]{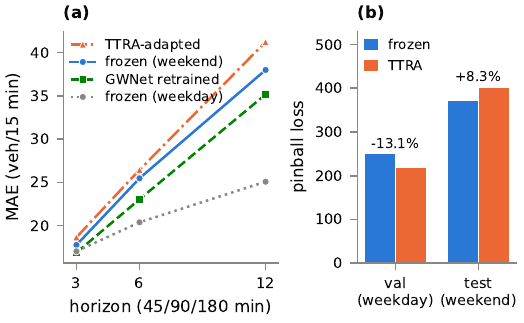}}
\caption{Test-time adaptation on LargeST-SD weekends. (a) The adapted
model has higher MAE at every horizon. (b) The adapter reduces weekday
validation pinball loss by $13.1\%$ but increases weekend pinball loss
by $8.3\%$.}
\label{fig:case}
\end{figure}

Retraining shows that exposure to the weekend regime matters. We
therefore ask whether the frozen model can recover from the test
context alone, without weekend targets. We use TTRA, a pre-registered
adapter that adds fast weights~\cite{vit3} to the frozen backbone and updates them by one self-supervised gradient
step on each context window, with no future target entering the
update. Its initialization is meta-learned on weekday windows using
the backbone's pinball loss~\cite{koenker1978},

\begin{equation}
\rho_\tau(y, \hat{q}) \;=\; 2\,\bigl|\,(y - \hat{q})\,
\bigl(\mathbf{1}\{y \le \hat{q}\} - \tau\bigr)\bigr|,
\label{eq:pinball}
\end{equation}

averaged over the forecast horizon and the backbone's 21 quantile
levels. On weekday validation, the adapter reduces pinball loss by 13.1\%.
On weekends, however, the same procedure increases pinball loss by
8.3\% and MAE from 26.14 to 27.80.
The adapted model has higher MAE at every forecast horizon (Fig.~\ref{fig:case}). TTRA
therefore improves its weekday objective without transferring that
improvement to the failing regime.
A shorter context does not repair the failure either
(Table~\ref{tab:retrain}). With 48 hours of context instead of 128,
weekend MAE changes by about 3\% and stays above the historical
average; with 24 hours it rises, and weekday MAE more than doubles for
both models.

The TTRA result leaves a second model-level possibility. Our mechanism
analysis shows that per-window normalization removes level and scale
changes before forecasting, so we test whether giving Graph WaveNet
the same normalization makes the weekend shift easier to absorb. The
matched variant replaces its training-set normalization constants with
each window's own per-sensor statistics, placing the estimator in the
form of Proposition~1 while leaving everything else unchanged. The
relative weekday-to-weekend gap does narrow, from 43\% to 26\% at the
pinned seed, but the model reaches $27.85 \pm 0.61$ MAE. The
normalization changes the relative degradation without recovering the
lost accuracy.

None of these interventions recovered the weekend failure. Our 
weekday retraining never sees the failing regime, our TTRA adapter 
improves its weekday objective while worsening weekend performance, 
and our matched-normalization variant narrows the relative gap 
without recovering the lost accuracy. Each of them tries to make a single model handle both regimes. 
We instead use a fact
these model-level repairs do not exploit: whether a window falls on a
weekday or weekend is known from its timestamp before forecasting.
This lets us select a model for the regime rather than require one
model to handle both. We therefore
route by timestamp, choosing among the historical average, 
Chronos-2, and TimesFM by validation MAE within each regime, which
requires labeled validation data from each regime. The
rule selects Chronos-2 for weekdays and the historical average for 
weekends. On the frozen test period it reaches 21.57 MAE on weekends 
and 20.46 over the full period, improving on both frozen TSFMs and 
on DCRNN's 21.20, though standard-trained Graph WaveNet remains 
better at 18.50 (Table~\ref{tab:tbig}). We select entirely on 
validation data and never touch the test set. 
Our repair experiments therefore mark a boundary. Models trained
with weekend data handle the regime, while weekday-only retraining,
test-time adaptation, and matched normalization do not recover the
failure. For the frozen deployment, routing works by using the known
regime to select the model that performed best on validation.

\subsection{Pre-Registered Predictions vs.\ Outcomes}
\label{sec:preds}

Table~\ref{tab:preds} reports each pre-registered prediction against
its outcome, including the two falsifications. We registered the
predictions on 2026-06-08 and amended P4 to P4$'$ on 2026-07-04, before
any adapter was trained. P4 originally predicted that the adapter on
frozen Chronos-2 would cut shift-induced degradation by at least 10\%
on two of the three shifts. The frozen-degradation grid falsified its
premise before any adapter existed, since Chronos-2 does not degrade on
most settings, and P4$'$ retargets the one cell where degradation is
observed on both backbones. The only substantial ranking reordering occurs on LargeST-SD, where
STAEformer was not part of the registered ranking set. 

\begin{table}[htbp]
\centering
\renewcommand{\arraystretch}{1.15}
\footnotesize
\caption{Pre-registered predictions and their outcomes.}
\label{tab:preds}
\begin{tabular}{@{}p{0.16in}>{\raggedright\arraybackslash}p{1.00in}>{\raggedright\arraybackslash}p{1.12in}p{0.52in}@{}}
\toprule
\textbf{ID} & \textbf{Prediction} & \textbf{Outcome} & \textbf{Verdict} \\
\midrule
P1 & Ranking $\tau$ lower under S1 than S2 & Held on LargeST-SD ($0.333 < 0.867$); ties elsewhere & Partially held \\ \addlinespace[2pt]
P2 & Graph models degrade more than per-sensor models under S3 & Needs S3-retrained graph baselines; scoped out by protocol & Not testable \\ \addlinespace[2pt]
P3 & STAEformer top-2 under S2, drops $\geq 2$ ranks under S1 & Rank 1 under both shifts on every dataset evaluated & Falsified \\ \addlinespace[2pt]
P4$'$ & TTRA cuts frozen SD-S1 degradation by $\geq 10\%$ & Adapted MAE 27.80 $>$ frozen 26.14 at every horizon & Falsified \\ \addlinespace[2pt]
P5 & Inner-loop steps $> 0$ beat static registers & Not run; gated on the failed adaptation gate (Sec.~\ref{sec:case}) & Not evaluated \\
\bottomrule
\end{tabular}
\end{table}

\section{Sensitivity Analysis}\label{sec:ablation}
\begin{figure}[t]
\centerline{\includegraphics[width=\columnwidth]{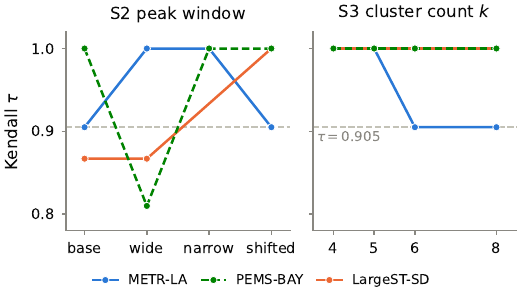}}
\caption{Ranking stability under other split choices: S2 peak windows
(left) and S3 cluster counts $k$ (right). Dashed line: $\tau=0.905$.}
\label{fig:split}
\end{figure}

We rerun the audit under three alternative peak-window definitions,
narrowing, widening, or shifting the registered Section~III-B window
by one hour, and under holdout cluster counts
$k \in \{4,5,6,8\}$. The off-peak improvement holds in all eleven
window--dataset pairs that admit a valid pool, and ranking
stability remains at $\tau \geq 0.81$ across the peak variants and
$\tau \geq 0.905$ across the cluster counts
(Fig.~\ref{fig:split}). The exception is the LargeST-SD spatial
result, whose degradation sign changes with the holdout definition.
This changes the non-degrading count from 7 of 9 to 6 of 9 but does
not affect the temporal failure or the severity-ranking result.

\section{Discussion}\label{sec:discussion}

\paragraph{\textbf{Regime-Level Monitoring}} These findings suggest that
monitoring a single severity score may not be enough for frozen TSFMs,
because the degrading regime can appear first as a change in model
ranking rather than as a large distribution shift. On LargeST-SD
weekends, this change reduces Kendall $\tau$ to 0.333 and moves the
historical average above both TSFMs
(Table~\ref{tab:tau}, Fig.~\ref{fig:slopes}). Yet the same weekend
setting receives the lowest severity reading in its dataset
(Table~\ref{tab1}). This mismatch is consistent with earlier traffic
studies showing that aggregate evaluation can conceal errors at finer
levels: corridor-level validation conceals per-lane errors
\cite{bishi_annsim26}, while aggregate MAE conceals
transition-regime error \cite{wangregime}. Our results show the same
problem at the shift level, where a single severity reading does not
identify the regime carrying the degradation
(Table~\ref{tab:deg}). This suggests that monitoring should also
compare model performance across known operating regimes. In the
weekend regime, the timestamp itself provides useful regime
information, allowing a historical average to outperform both TSFMs. The router in Section~\ref{sec:repair} uses this
information to select the better of the available frozen candidates in
that regime.

\paragraph{\textbf{Boundary of Distance-Based Severity}} Our findings do not
imply that Wasserstein distance is generally uninformative. Chen et al.
report correlations above 0.60 between Wasserstein distance and
foundation-model loss across datasets \cite{chenwass}. That comparison
is made across separate datasets, where we expect level and scale
differences to contribute strongly to the distance. Our setting is
different. Within one deployment domain, Proposition~1 shows that an
instance-normalizing forecaster removes the affine component that can
dominate marginal $W_1$. This removal can break the expected relationship
between the severity reading and model degradation
(Fig.~\ref{fig:scatter}). The accuracy-on-the-line result similarly
reports strong correlations across architectures, hyperparameters, and
training scale, while also identifying cases where that relationship
weakens \cite{aotl}. Regime shifts under instance normalization
provide another such case, as our experiments show. These
results suggest that a useful
severity measure should depend on the forecasting task, because a
distance on raw observable marginals can be dominated by information
that the model removes before prediction. Vision reports a similar
pattern, where behavior-based estimators predict degradation in
settings where distributional distances fail \cite{doc,atc}.
Quantile-spread and reference-residual monitors are possible
time-series alternatives, although this audit does not evaluate them.

\paragraph{\textbf{Normalization and Model-Class Differences}}
The different behavior of the two model classes is consistent with
their normalization schemes. The graph baselines normalize with
training-split statistics, while the frozen TSFMs re-standardize every
context window (Section~\ref{sec:decouple}). As a result, a regime whose
level differs from the training statistics reaches the graph models
without per-window normalization, while the frozen TSFMs remove that
level difference before prediction. This difference helps explain why
the same shift can affect the two model classes differently.
Section~\ref{sec:normlineage} traces these normalization designs and
their connection to severity-based auditing.

\paragraph{\textbf{Contamination and Within-Dataset Auditing}} These findings
must also be interpreted in light of possible training-data
contamination, since the backbones may have seen these datasets in
pretraining. PEMS-BAY and LargeST appear verbatim in at least one
public TSFM pretraining corpus, the one released with
Moirai~\cite{moirai}. The pretraining corpora of the two backbones we
evaluate are not public, so their exposure cannot be checked directly.
A recent audit finds only
a small fraction of benchmark datasets absent from every pretraining
corpus \cite{meyerleakage}, and methods for auditing this contamination
are emerging \cite{tsfmaudit}. Our within-dataset comparison reduces
this concern because the reference and shifted regimes come from the
same dataset. Any memorization can therefore affect performance on
both regimes rather than only one side of the comparison. The weekend
failure remains present under this comparison, regardless of the
backbones' possible exposure to the dataset.

\section{Limitations}
All evidence here comes from traffic data and two backbones, so it does
not show that severity fails for every instance-normalized forecaster
or in other domains. Nor do we show that every notion of severity
fails. A severity measure tied to the forecasting task remains open.

\section{Conclusion}
In this work, we conducted a systematic audit that shows that the tested measures used to rank distribution shifts by severity do not predict degradation for instance-normalized TSFMs in these settings. Our results show that the regime carrying its dataset's lowest severity score causes the largest degradation, and this degradation reorders the audited models' ranking, with a plain historical average model passing both Chronos-2 and TimesFM. We show that the equivariance property of these models provides a mechanism by which severity measures can fail to track degradation. Experiments across three datasets (METR-LA, PEMS-BAY, and LargeST-SD) on nine shifts show that severity measurement does not predict failure for instance-normalized TSFMs, because none of the severity measures we tested identifies the regimes in which these models fail. When these models fail, they fail without warning, and the failing regime has the lowest severity reading in its dataset with degradation of up to $29\%$ for these models. Lastly, test-time adaptation did not repair the failure, while models trained with weekend data did.

\section*{Acknowledgment}
Claude was used to assist with related-work synthesis and as a secondary check of the proof of Proposition~1. Claude also assisted with the coding. All results and claims
are the true work of the authors.


\begin{thebibliography}{00}

\bibitem{chronos} A. F. Ansari, L. Stella, C. Turkmen, et al., ``Chronos: learning the language of time series,'' \emph{Trans. Mach. Learn. Res.}, 2024.
\bibitem{timesfm} A. Das, W. Kong, R. Sen, and Y. Zhou, ``A decoder-only foundation model for time-series forecasting,'' in \emph{Proc. ICML}, 2024.
\bibitem{moirai} G. Woo, C. Liu, A. Kumar, C. Xiong, S. Savarese, and D. Sahoo, ``Unified training of universal time series forecasting transformers,'' in \emph{Proc. ICML}, 2024.
\bibitem{moment} M. Goswami, K. Szafer, A. Choudhry, Y. Cai, S. Li, and A. Dubrawski, ``MOMENT: a family of open time-series foundation models,'' in \emph{Proc. ICML}, 2024.
\bibitem{chronos2} A. F. Ansari, O. Shchur, J. K\"uken, et al., ``Chronos-2: from univariate to universal forecasting,'' arXiv:2510.15821, 2025.
\bibitem{dcrnn} Y. Li, R. Yu, C. Shahabi, and Y. Liu, ``Diffusion convolutional recurrent neural network: data-driven traffic forecasting,'' in \emph{Proc. ICLR}, 2018.
\bibitem{gwnet} Z. Wu, S. Pan, G. Long, J. Jiang, and C. Zhang, ``Graph WaveNet for deep spatial-temporal graph modeling,'' in \emph{Proc. IJCAI}, 2019.
\bibitem{staeformer} H. Liu, Z. Dong, R. Jiang, et al., ``Spatio-temporal adaptive embedding makes vanilla transformer SOTA for traffic forecasting,'' in \emph{Proc. CIKM}, 2023.
\bibitem{largest} X. Liu, Y. Xia, Y. Liang, et al., ``LargeST: a benchmark dataset for large-scale traffic forecasting,'' in \emph{Proc. NeurIPS Datasets and Benchmarks}, 2023.
\bibitem{gama2014} J. Gama, I. \v{Z}liobait\.e, A. Bifet, M. Pechenizkiy, and A. Bouchachia, ``A survey on concept drift adaptation,'' \emph{ACM Comput. Surv.}, vol. 46, no. 4, 2014.
\bibitem{drifttools} R. M\"uller, M. Abdelaal, and D. Stjelja, ``Open-source drift detection tools in action: insights from two use cases,'' in \emph{Proc. Int. Conf. Big Data Analytics and Knowledge Discovery (DaWaK)}, 2024.
\bibitem{driftdlrep} S. Greco, B. Vacchetti, D. Apiletti, and T. Cerquitelli, ``Unsupervised concept drift detection from deep learning representations in real-time,'' \emph{IEEE Trans. Knowl. Data Eng.}, vol. 37, no. 10, pp. 6232--6245, 2025.
\bibitem{driftbench} D. Soukup, R. Pln\'y, D. Va\v{s}ata, and T. \v{C}ejka, ``Drift-based dataset stability benchmark,'' arXiv:2512.23762, 2025.
\bibitem{peyrecuturi} G. Peyr\'e and M. Cuturi, ``Computational optimal transport,'' \emph{Found. Trends Mach. Learn.}, vol. 11, no. 5--6, pp. 355--607, 2019.
\bibitem{koenker1978} R. Koenker and G. Bassett, ``Regression quantiles,'' \emph{Econometrica}, vol. 46, no. 1, pp. 33--50, 1978.
\bibitem{failingloudly} S. Rabanser, S. G\"unnemann, and Z. C. Lipton, ``Failing loudly: an empirical study of methods for detecting dataset shift,'' in \emph{Proc. NeurIPS}, 2019.
\bibitem{uncertaintyshift} Y. Ovadia, E. Fertig, J. Ren, et al., ``Can you trust your model's uncertainty? Evaluating predictive uncertainty under dataset shift,'' in \emph{Proc. NeurIPS}, 2019.
\bibitem{wilds} P. W. Koh, S. Sagawa, H. Marklund, et al., ``WILDS: a benchmark of in-the-wild distribution shifts,'' in \emph{Proc. ICML}, 2021.
\bibitem{revin} T. Kim, J. Kim, Y. Tae, C. Park, J.-H. Choi, and J. Choo, ``Reversible instance normalization for accurate time-series forecasting against distribution shift,'' in \emph{Proc. ICLR}, 2022.
\bibitem{nonstattransf} Y. Liu, H. Wu, J. Wang, and M. Long, ``Non-stationary transformers: exploring the stationarity in time series forecasting,'' in \emph{Proc. NeurIPS}, 2022.
\bibitem{dishts} W. Fan, P. Wang, D. Wang, D. Wang, Y. Zhou, and Y. Fu, ``Dish-TS: a general paradigm for alleviating distribution shift in time series forecasting,'' in \emph{Proc. AAAI}, 2023.
\bibitem{tttsun} Y. Sun, X. Wang, Z. Liu, J. Miller, A. A. Efros, and M. Hardt, ``Test-time training with self-supervision for generalization under distribution shifts,'' in \emph{Proc. ICML}, 2020.
\bibitem{vit3} D. Han, Y. Li, T. Li, Z. Cao, Z. Wang, J. Song, Y. Cheng, B. Zheng, and G. Huang, ``ViT$^3$: unlocking test-time training in vision,'' arXiv:2512.01643, 2025.
\bibitem{princtta} H. Wang, R. Xu, G. Kementzidis, K. Cho, S. Ramirez Villarreal, and Y. Deng, ``Towards principled test-time adaptation for time series forecasting,'' arXiv:2605.17250, 2026.
\bibitem{peft_tta} H. R. Medeiros, H. Sharifi-Noghabi, G. L. Oliveira, and S. Irandoust, ``Accurate parameter-efficient test-time adaptation for time series forecasting,'' in \emph{ICML Workshop on Test-Time Adaptation}, 2025. arXiv:2506.23424.
\bibitem{onlinefeatadj} X. Huang, S. Qiu, J. Du, and C. Yang, ``Online time series prediction using feature adjustment,'' in \emph{Proc. ICLR}, 2026.
\bibitem{tsfmaudit} H. Li, S. Xie, L. Shen, et al., ``TSFMAudit: data contamination auditing in forecasting time series foundation models,'' arXiv:2605.26161, 2026.
\bibitem{tsfmpower} M. E. Za'ter and B.-M. Hodge, ``Empirical assessment of time-series foundation models for power system forecasting applications,'' arXiv:2604.22077, 2026.
\bibitem{yanespulido} J. Yanes-Pulido and F. Rodrigues, ``Time series foundation models as strong baselines in transportation forecasting: a large-scale benchmark analysis,'' arXiv:2602.24238, 2026.
\bibitem{tonercloud} W. Toner, T. L. Lee, A. Joosen, R. Singh, and M. Asenov, ``Performance of zero-shot time series foundation models on cloud data,'' in \emph{ICLR Workshop ``I Can't Believe It's Not Better''}, 2025. arXiv:2502.12944.
\bibitem{maggio} S. Maggio, V. Bouvier, and L. Dreyfus-Schmidt, ``Performance prediction under dataset shift,'' in \emph{Proc. ICPR}, 2022.
\bibitem{woods} J.-C. Gagnon-Audet, K. Ahuja, M.-J. Darvishi-Bayazi, P. Mousavi, G. Dumas, and I. Rish, ``WOODS: benchmarks for out-of-distribution generalization in time series,'' \emph{Trans. Mach. Learn. Res.}, 2023.
\bibitem{bishi_annsim26} K. R. Bishi, C. Bowman, and J. A. Miller, ``Beyond corridor averages: lane-level validation of microscopic freeway simulation with empirically parameterized arrivals,'' in \emph{Proc. Annual Modeling and Simulation Conf. (ANNSIM)}, SCS, 2026.
\bibitem{wangregime} Y. Wang, X. Sun, L. Kong, W. Gao, Y. Li, Z. Fan, and Z. Zhuang, ``Do time series foundation model benchmarks hide regime-dependent failures? Evidence from traffic speed forecasting,'' in \emph{ICML Workshop on Forecasting as a New Frontier of Intelligence}, 2026. arXiv:2606.18367.
\bibitem{chenwass} H. Chen, A. Mehra, J. Kimball, and R. A. Rossi, ``Measuring time-series dataset similarity using Wasserstein distance,'' arXiv:2507.22189, 2025.
\bibitem{aotl} J. Miller, R. Taori, A. Raghunathan, S. Sagawa, P. W. Koh, V. Shankar, P. Liang, Y. Carmon, and L. Schmidt, ``Accuracy on the line: on the strong correlation between out-of-distribution and in-distribution generalization,'' in \emph{Proc. ICML}, 2021.
\bibitem{doc} D. Guillory, V. Shankar, S. Ebrahimi, T. Darrell, and L. Schmidt, ``Predicting with confidence on unseen distributions,'' in \emph{Proc. ICCV}, 2021.
\bibitem{atc} S. Garg, S. Balakrishnan, Z. C. Lipton, B. Neyshabur, and H. Sedghi, ``Leveraging unlabeled data to predict out-of-distribution performance,'' in \emph{Proc. ICLR}, 2022.
\bibitem{meyerleakage} M. Meyer, S. Kaltenpoth, K. Zalipski, and O. M\"uller, ``Rethinking evaluation in the era of time series foundation models: (un)known information leakage challenges,'' arXiv:2510.13654, 2025.
\bibitem{normequiv23} S. Herbreteau, E. Moebel, and C. Kervrann, ``Normalization-equivariant neural networks with application to image denoising,'' in \emph{Proc. NeurIPS}, 2023.
\bibitem{normequiv26} Y. Saied and F. Fleuret, ``Normalization equivariance for arbitrary backbones, with application to image denoising,'' in \emph{Proc. ICML}, PMLR vol. 306, 2026.
\bibitem{rdumb} O. Press, S. Schneider, M. K\"ummerer, and M. Bethge, ``RDumb: a simple approach that questions our progress in continual test-time adaptation,'' in \emph{Proc. NeurIPS}, 2023.
\bibitem{note_tta} T. Gong, J. Jeong, T. Kim, Y. Kim, J. Shin, and S.-J. Lee, ``NOTE: robust continual test-time adaptation against temporal correlation,'' in \emph{Proc. NeurIPS}, 2022.
\bibitem{ttapitfalls} H. Zhao, Y. Liu, A. Alahi, and T. Lin, ``On pitfalls of test-time adaptation,'' in \emph{Proc. ICML}, 2023.
\end{thebibliography}
\end{document}